# Zur Darstellung eines mehrstufigen Prototypbegriffs in der multilingualen automatischen Sprachgenerierung: vom Korpus über *word embeddings* bis hin zum automatischen Wörterbuch

María José Domínguez Vázquez, *Universidade de Santiago de Compostela — ILG, Spain (majo.dominguez@usc.es)*

**Zusammenfassung:** Das multilinguale Wörterbuch zur Substantivvalenz *Portlex* gilt als Ausgangspunkt für die Entstehung der automatischen Sprachgeneratoren *Xera* und *Combinatoria,* deren Entwicklung und Handhabung hier präsentiert wird. Beide Prototypen dienen zur automatischen Generierung von Nominalphrasen mit ihren mono- und biargumentalen Valenzstellen, die u.a. als Wörterbuchbeispiele oder als integrierte Bestandteile künftiger autonomer E-Learning-Tools eine Anwendung finden könnten. Als Modelle für neuartige automatische Valenzwörterbücher mit Benutzerinteraktion fassen wir die Sprachgeneratoren in ihrem heutigen Zustand auf.

Bei dem spezifischen methodologischen Verfahren zur Entwicklung der Sprachgeneratoren stellt sich die syntaktisch-semantische Beschreibung der vom Valenzträger eröffneten Leerstellen aus syntagmatischer und paradigmatischer Sicht als Schwerpunkt heraus. Zusammen mit Faktoren wie der Repräsentativität, der grammatischen Korrektheit, der semantischen Kohärenz, der Frequenz und der Vielfältigkeit der lexikalischen Kandidaten sowie der semantischen Klassen und der Argumentstrukturen, die feste Bestandteile beider Ressourcen sind, sticht ein mehrschichtiger Prototypsbegriff hervor. Die kombinierte Anwendung dieses Prototypbegriffs sowie von *word embeddings* zeigt zusammen mit Techniken aus dem Gebiet der maschinellen Verarbeitung und Generation natürlicher Sprache (NLP und NLG) einen neuen Weg zur künftigen Entwicklung von automatisch generierten plurilingualen Valenzwörterbüchern.

Insgesamt stellt der Beitrag die Sprachgeneratoren sowohl aus der Perspektive ihrer Entwicklung als auch aus Nutzersicht dar. Der Fokuss wird auf die Rolle des Prototypsbegriffs bei der Entwicklung der Ressourcen gelegt.

**Schlüsselwörter:** NLG: *NATURAL LANGUAGE GENERATION*, AUTOMATISCHES WÖRTERBUCH, INTERAKTIVES WÖRTERBUCH, SPRACHGENERATOREN, KORPUSLEXIKOGRAPHIE, ONTOLOGIE, PROTOTYP, LEXIKALISCHER PROTOTYP, SEMANTISCHE PROTOTYPISCHE KLASSEN

**Abstract: Towards the Description of a Multi-sided Prototype Concept in Multilingual Automatic Language Generation: From Corpus via *Word***

***Embeddings* to the Automatic Dictionary.** The multilingual dictionary of noun valency *Portlex* is considered to be the trigger for the creation of the automatic language generators *Xera* and *Combinatoria*, whose development and use is presented in this paper. Both prototypes are used for the automatic generation of nominal phrases with their mono- and bi-argumental valence slots, which could be used, among others, as dictionary examples or as integrated components of future autonomous E-Learning-Tools. As samples for new types of automatic valency dictionaries including user interaction, we consider the language generators as we know them today.

In the specific methodological procedure for the development of the language generators, the syntactic-semantic description of the noun slots turns out to be the main focus from a syntagmatic and paradigmatic point of view. Along with factors such as representativeness, grammatical correctness, semantic coherence, frequency and the variety of lexical candidates, as well as semantic classes and argument structures, which are fixed components of both resources, a concept of a multi-sided prototype stands out. The combined application of this prototype concept as well as of *word embeddings* together with techniques from the field of automatic natural language processing and generation (NLP and NLG) opens up a new way for the future development of automatically generated plurilingual valency dictionaries.

All things considered, the paper depicts the language generators both from the point of view of their development as well as from that of the users. The focus lies on the role of the prototype concept within the development of the resources.

## 1.    Einführung

Der Ausgangspunkt für die Gestaltung der Sprachgeneratoren *Xera* und *Combinatoria* (Domínguez Vázquez et al. 2020) ist das online Wörterbuch zur Substantivvalenz *Portlex* (Domínguez Vázquez et al. 2018). *Portlex* ist ein semikollaboratives, multilinguales, *cross-lingual* Wörterbuch für Deutsch, Spanisch, Französisch, Italienisch und Galicisch (Domínguez Vázquez und Valcárcel Riveiro 2020). Bei seiner Entwicklung haben wir feststellen können, dass die Erhebung von Korporabelegen[1] zur Veranschaulichung der vom Substantiv eröffneten Valenzstellen sowie insbesondere zur Beschreibung des Kombinationspotentials der verschiedenen Ausdrucksformen jedes Arguments für fünf Sprachen viel Zeit in Anspruch nimmt. Besonders bei Suchanfragen zur Verdeutlichung der Interaktion der Ergänzungen hat sich eine nicht zu unterschätzende Anzahl an erhobenen Korporabelegen angesichts des Wörterbuchtyps nicht immer als adäquat erwiesen. Das betrifft vor allem die Realisierung der Ergänzungen in Form eines Adjektivs oder eines Kompositabestandteils, denn — wie allgemein bekannt — geht die Korpushäufigkeit einer Ausdrucksform nicht unbedingt mit ihrem Ergänzungsstatus einher.

Aufgrund der Tatsache, dass die verwendeten Korpora zur Anfertigung von *Portlex* weder funktional-syntaktisch noch semantisch annotiert sind[2] —

http://lexikos.journals.ac.za; https://doi.org/10.5788/31-1-1623

Vom Korpus über *word embeddings* bis hin zum automatischen Wörterbuch 3

dementsprechend lassen sich die Korporabelege und -daten nicht nach diesen Parametern abrufen —, haben wir uns dazu entschieden, die Perspektive umzukehren: anstatt Korporabelege zur Darstellung der Nominalvalenz anzubieten, korpusgestützte Beispiele für die jeweiligen Valenzmuster automatisch zu generieren. So sind *Xera* und *Combinatoria* entstanden.

Beide Prototypen vermitteln automatisch generierte Nominalphrasen für jeweils 20 Substantive im Spanischen, Deutschen und Französischen[3]. *Xera* erzeugt einfache syntaktisch-semantische Argumentstrukturen — wie z.B. *der Geruch nach Blumen*; *Combinatoria* trägt zur Vermittlung komplexerer nominaler Valenzmuster bei, wie z.B *der Geruch des Zimmers nach Blumen* (siehe 3.). Ferner sind die Generatoren zur Einbindung weiterer Beschreibungseinheiten sowie Sprachen gestaltet und lassen sich ebenfalls in weitere Ressourcen integrieren und dadurch wieder verwerten.

In beiden Ressourcen ist eine valenzfundierte Beschreibung des nominalen Kombinationspotentials mit Fokus auf die kombinatorische Bedeutung (Engel 1996, 2004) von unentbehrlicher Bedeutung. Es geht hierbei nicht nur um die Frage, ob bei der Wiedergabe einer semantischen Rolle eine bestimmte ontologische Entität eine Valenzstelle belegen kann (oder nicht), sondern auch darum, welche konkreten lexikalischen Kandidaten bzw. ontologischen Kategorien diese Stelle einnehmen können. Diese Herangehensweise hängt mit dem Hauptziel von Valenzwörterbüchern zusammen, die Schumacher (2006: 1396) so beschreibt:

> Ein Defizit dieser [grammatikographischen] Darstellungsform besteht darin, dass hierbei die Regeln immer nur mit relativ wenigen Beispielen illustriert werden können. Dadurch bleibt für den Benutzer die Frage bestehen, welche der Regeln für die vielen Fälle zutreffen könnten, die nicht als grammatikalisches Beispiel dienen. Hier kann ein Lexikon Abhilfe schaffen [...].

Bei der Anfertigung der Sprachgeneratoren hat sich die methodologische Frage, wie man eine semantischfundierte Beschreibung der nominalen Valenzstellen automatisch erlangt und generiert, als eine Herausforderung erwiesen. Zu ihrer Beantwortung sind wir von der Anwendung einer „kombinierten Methode" mit Rückgriff auf die Interoperabilität von Ressourcen ausgegangen. Im Konkreten baut das methodologische Verfahren zur Festlegung der anzubietenden nominalen Argumentstrukturen sowie der Beispiele auf (a) einer valenzfundierten Beschreibung des Substantivs und der zu belegenden Leerstellen, (b) einer korpusgestützten Analyse erhobener Ko-Okurrenzen sowie ihrem Kombinationspotential, (c) eine automatische Datenerhebung von lexikalischen Daten aus NLP-Ressourcen und aus den Wordnets für die beschriebenen Sprachen, vorhanden in dem *Multilingual Central Repository*[4] (s. Domínguez Vázquez, Solla Portela und Valcárcel Riveiro 2019) und (d) einem mehrschichtigen Prototypbegriff auf.

Prototypen sind in unserem Ansatz typische bzw. repräsentative Instanzen für eine konkrete Slotbesetzung oder für eine konkrete Argumentstruktur. Der Beschreibung des Prototypbegriffs weist dieser Beitrag einen gesonderten

Stellenwert zu, denn er gilt als Verknüpfungselement bei der automatischen Datenerhebung und -generierung. Weiterhin sind die von Generatoren gelieferten Beispiele und Muster im Hinblick auf ihre Repräsentativität, semantische Kompabilität sowie grammatische Korrektheit, Vielfältigkeit und Prototypizität sowohl manuell als auch automatisch herausgefiltert worden[5]:

— Es ist allgemein bekannt, dass grammatische Korrektheit mit semantischer Akzeptabilität nicht immer einhergeht, sowie dass eine Ausdrucksform mit unterschiedlichen Bedeutungen korrelieren kann. Daher wird die syntaktische und semantische Analyse der vom Valenzträger geforderten Argumenten, ihrer Interaktions- sowie (In)kompatibilitätsregeln und ihrer Distribution unter Berücksichtigung valenzfundierter Kriterien sowie der Korrektheit und Akzeptabilität zum Schlüsselkonzept. Es handelt sich hier um die Analyse der kombinatorischen Bedeutung im Sinne von Engel (2004: 188).

— Zur Gewährleistung repräsentativer und prototypischer Beispiele werden zum einen Belege aus den Korpora aus *Sketch Engine* herangezogen[6] und nach Häufigkeitskriterien sortiert, zum anderen bedarf die Entfaltung des im Nomen enthaltenen Potentials einer näheren Untersuchung, die mit Rückgriff auf den Prototypbegriff sowie auf die Anwendung der prädiktiven Methode *word2vec (word embeddings)* erfolgt.

— Die semantische Zuordnung des Sprachmaterials und die Etablierung von semantischen Klassen wird mithilfe der Ontologien aus WordNet[7] erlangt. Dazu greift man ebenfalls auf die Prototypizität zurück.

Der Fokus des Beitrags liegt somit auf der Darstellung des methodologischen Verfahrens und des den Generatoren zugrundeliegenden mehrschichtigen Prototypbegriffs, d.h. auf der Entwicklung der Generatoren. Das ist u.E. insofern relevant, dass sie ein computergestütztes Modell für die Erstellung von automatisch generierten plurilingualen Valenzwörterbüchern vorstellen. Einblicke in die Nutzerperspektive lassen sich auch gewinnen.

Der Beitrag ist wie folgt aufgebaut: Im Abschnitt 2 wird ein Gesamtüberblick über Grundeigenschaften der Generatoren gegeben. Abschnitt 3 verdeutlicht angesichts eines mehrschichtigen Prototypbegriffs das methodologische Verfahren. Auf die Wechselwirkung zwischen Typizität, Frequenz und Repräsentativität bei der Auswahl der Argumentstrukturen, lexikalischer Kandidaten und semantischer Klassen wird in 4 eingegangen. Der 5. Abschnitt setzt sich mit den Generatoren als Modell für ein automatisches plurilinguales Wörterbuch auseinander.

## 2.    Allgemeines zur Typologie der multilingualen Sprachgeneratoren

Das wissenschaftliche Interesse an der natürlichsprachlichen Generierung (NLG) ist seit den 90er Jahren gewachsen. Heutzutage gibt es unterschiedliche auto-matische Generatoren natürlicher Sprache, deren Ziel die Erzeugung menschen-ähnlicher Texte jeglicher Art ist. Als Beispiele lassen sich meteorologische oder sportliche Berichte, medizinische Zusammenfassungen, vereinfachte Texte, Empfehlungstexte und Dialoge, u.a. (Vicente et al. 2015, Nallapati et al. 2016, Sordoni et al. 2015) anführen. Ebenfalls ist es möglich, aus Bildern Texte auto-matisch zu generieren und umgekehrt (Otter et al. 2020) sowie mit einer geringen Menge an Ausgangsmaterial Witze, Gedichte und Geschichten (Roemmele 2016) zu erstellen. Auf lexikographischem Gebiet rechnet man auch mit Vorschlägen zur automatischen Generierung von Wörterbüchern (Bardanca Outeiriño 2020, Héja und Takács 2012, Kabashi 2018) sowie Wörterbucheinträgen (Geyken et al. 2017), zur Identifizierung von mikrostrukturellen Teilen, wie z.B. den Bei-spielen (Kilgarriff et al. 2008), oder zu ihrer automatischen Erhebung (Kosem et al. 2019). Das Endprodukt unserer Generatoren ist ein anderes, und zwar syn-taktisch-semantisch akzeptable einfache und komplexe Nominalphrasen im Kontext[8]. In beiden Generatoren handelt es sich nämlich um die Vermittlung einer ausreichenden und distinktiven Auskunft über den vom Valenzträger festgelegten syntaktisch-semantischen Rahmen, d.h. über die syntaktisch-semantische Schnittstelle der Nominalphrase.

## 3. Allgemeine Beschreibung der Sprachgeneratoren *Xera* und *Combinatoria*

### 3.1 Beschreibungsebenen: die Festlegung der Argumentstruktur

Jeweils 20 Substantive des Deutschen, Spanischen und Französischen, deren Auswahl auf ihre Einordnung in unterschiedliche Szenen zurückgeht, sind zur Auswertung der Prototypen herangezogen worden (Tab. 1).

| Szenen | Substantive |
|---|---|
| BEWEGUNG | huida \| Flucht \| fuite<br>viaje \| Reise \| voyage<br>mudanza \| Umzug \| déménagement |
| LOKATION | presencia \| Anwesenheit \| présence<br>ausencia \| Abwesenheit \| absence<br>estancia \| Aufenthalt \| séjour |
| AUSDRUCK | conversación \| Gespräch \| conversation<br>discusión \| Diskussion \| discussion<br>pregunta \| Frage \| question<br>respuesta \| Antwort \| réponse<br>texto \|Text \| texte<br>video \| Video \| vidéo |

| AFFIZIERTHEIT | muerte \| Tod \| mort |
| | aumento \| Zunahme \|augmentation |
| | dolor \| Schmerz \| douleur |
| | amor \| Liebe \| amour |
| KLASSIFIKATION | olor \| Geruch \| odeur |
| | sabor \| Geschmack \| saveur |
| | color \| Farbe \| couleur |
| | anchura \| Breite \| largeur |

**Tab. 1:**    Substantive in *Xera* und *Combinatoria*

Für die Gewährleistung der automatischen Erhebung und Generierung sprachlicher Daten sind unterschiedliche Arbeitsphasen und -verfahren notwendig: (i) Festlegung der Argumentstruktur, (ii) semantische Prototypisierung, (iii) Expandierung der Prototypen, iv) Debugging und v) paradigmatische Verpackung und automatische Generierung der Nominalphrase (s. Domínguez Vázquez, Solla Portela und Valcárcel Riveiro 2019). Die unterschiedlichen Arbeitsschritte und die dafür eingesetzten Ressourcen und Tools fasst die Abb. 1[9] zusammen:

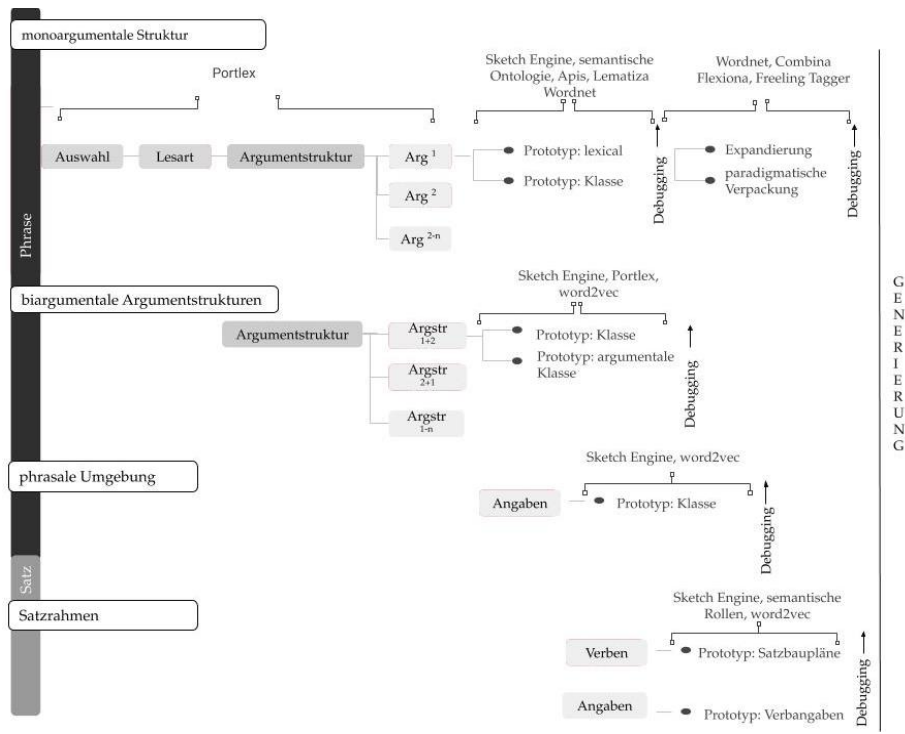

**Abb. 1:**    Beschreibungsebenen und Ressourcen

Die erste Beschreibungsphase[10] betrifft die Festlegung der Argumentstruktur, was am Beispiel des Substantivs TEXT veranschaulicht wird: Abb. 2[11] zeigt die Analyse der Valenzstellen, Abb. 3 einige biargumentale Muster.

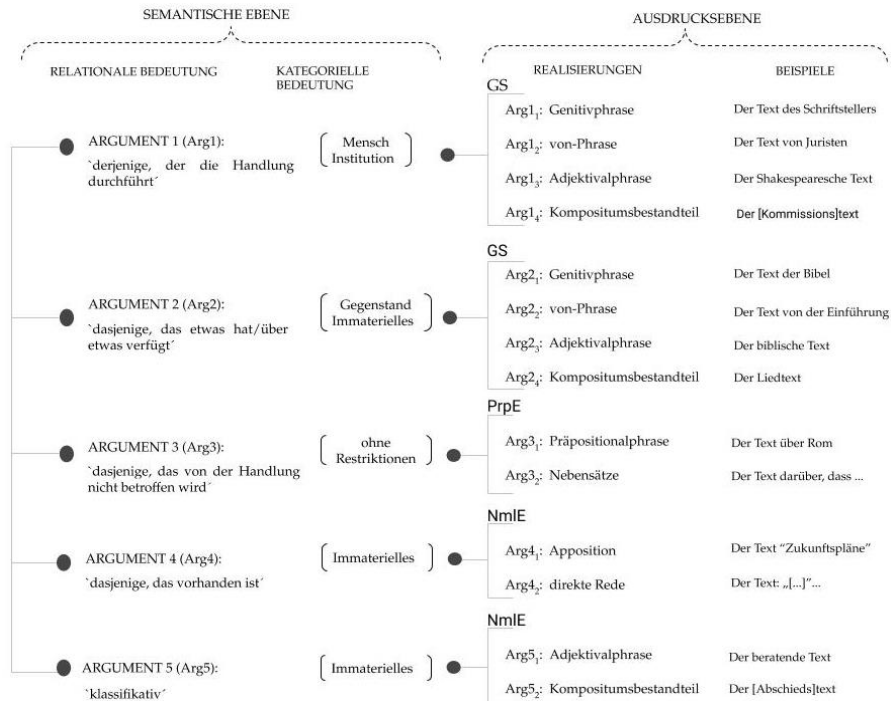

**Abb. 2:**     Argumentstellen von TEXT

| biargumentale Argumentstrukturen | |
|---|---|
| $Arg1_1 + Arg3_4$ | der Text einer Journalistin darüber, warum man in diesem Ungarn nicht mehr arbeiten kann, ... |
| $Arg1_4 + Arg3_1$ | der Regierungstext über das Gesetz |
| $Arg1_3 + Arg3_1$ | der Platonische Text über Parmenides |
| $Arg2_1 + Arg3_1$ | der Text der Bibel über die Versuchung… |
| $Arg5_2 + Arg3_1$ | der Predigttext über die Hochzeit zu Kana... |
| $Arg5_2 + Arg1_2$ | der Pressetext von Volkswagen... |
| $Arg5_2 + Arg4_1$ | der Abschiedstext "Cordials Salids" ... |
| $Arg5_1 + Arg4_2$ | der liturgische Text: "Dich, Gott, loben wir"... |

**Abb. 3:**     Biargumentale Muster von TEXT

Wie aus der Abb. 2. hervorgeht, steht in unserem Modell ein semantisch verankerter valenzfundierter Ansatz im Vordergrund:

— Die jeweiligen Argumentstellen werden nach ihrer relationalen Bedeutung — wie bei Arg1 `derjenige, der die Handlung durchführt´ — analysiert. Die Beschreibung der semantischen Rollen ermöglicht den notwendigen Unterschied zwischen einem $Arg1_4$, z.B. 'Kommissionstext', gegenüber einem $Arg5_2$ wie 'Abschiedstext'.

— Die Kandidaten zur Wiedergabe einer Rolle bzw. zur Besetzung einer bestimmten Leerstelle bestimmen wir ebenfalls kategoriell[12] (s. 4.3.) und schreiben sie einer semantischen Klasse (s. 4.4.) zu.

Das bereits Angeführte sollte nicht den Eindruck erwecken, dass die Ausdrucksebene außer Acht gelassen wird, ganz im Gegenteil: eine vollständige Darstellung valenzbezogener Argumentstrukturen setzt die Beschreibung möglicher Realisierungen sowie ihres Kombinationspotentials voraus, und dies für die drei beschriebenen Sprachen.

### 3.2    Nutzer und Nutzung der Ressourcen

FremdsprachenstudentInnen und -dozentInnen stellen sich als den an erster Stelle anvisierten Benutzerkreis heraus. Beide können bei unterschiedlichen Benutzungszielen und -situationen von den Ressourcen profitieren, vor allem wenn man in Betracht zieht, dass (i) die auf die Valenz zurückgehenden Fehler im Fremdsprachenunterricht häufig vorkommen (Gao und Liu 2020, Nied Curcio 2014) und (ii) das Erlernen des Wortschatzes eng mit der Aneignung der syntaktisch-semantischen Umgebung zusammenhängt (Laufer und Nation 2012). Als sekundäre Nutzer kommen Wissenschaftler und Computerlexika, die sprachliches Wissen benötigen, in Frage (vgl. 5).

Bei der Handhabung beider Tools ist vom menschlichen Nutzer zunächst die Sprache und das nachzuschlagende Substantiv auszuwählen. Anschließend entsteht in beiden Tools ein Dropdown-Menü: das Argumentstrukturmenü in *Xera* (Abb. 4) und eine zu selegierende Kombination von Argumentstellen in *Combinatoria* (Abb. 7).

**Abb. 4:**     Argumentstrukturmenü bei *Xera*. Beispiel TEXT

Je nach selegiertem Muster wird ein weiteres Menü mit den semantischen prototypischen Klassen (s. 4.4.) der ausgewählten Leerstelle entfaltet. Somit liefert *Xera* beim Muster [Determinant + Adjektiv + TEXT + Determinant im Genitiv + Arg1] die folgenden semantischen Klassen[13], begleitet von Standardbeispielen:

**Abb. 5:**     Auswahlmenü semantischer Klassen bei *Xera*. Beispiel TEXT

Selegiert man in *Xera* dann eine konkrete semantische Klasse, z.B. {ANIMADO HUMANO PROFESIÓN EDUCACIÓN}[14], werden Beispiele zum syntaktischen sowie semantisch relational-kategoriellen monoargumentalen Potential automatisch vermittelt:

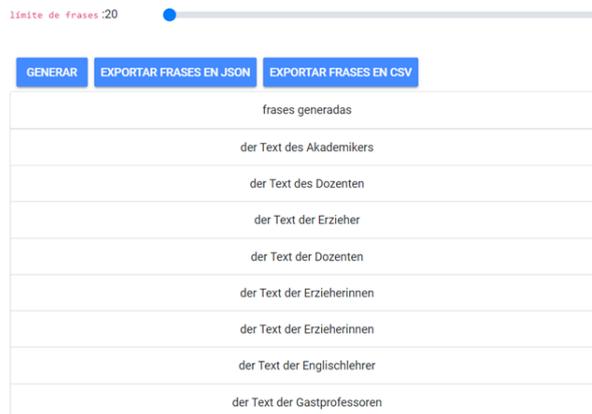

**Abb. 6:**    Beispiel für automatisch generierte Beispiele aus *Xera* für [DETERMI-NANT + TEXT + DETERMINANT IM GENITIV + ARG1₁]

Möchte man biargumentale Schemata suchen, steht *Combinatoria* zur Verfügung (Abb. 7)[15], die sich auf die zuvor programmierten Selektionsbeschränkungen sowie syntagmatischen Relationen stützt.

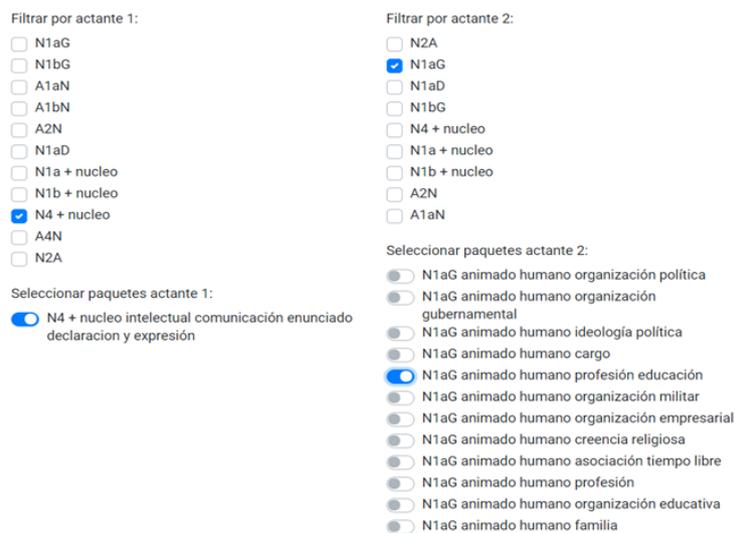

**Abb. 7:**    Argumentstrukturmenü bei *Combinatoria* 1.0. Beispiel TEXT

Gemäß der Auswahl (s. Abb. 7) erhält man beim Anklicken auf „Generar fra-
ses" automatisch generierte Beispiele, die sich im JSON- und CSV-Format
exportieren lassen:

[DETERMINANT + ARG5$_{2+}$ TEXT + DETERMINANT IM GENITIV + ARG1$_1$]

der Bemerkungstext der Akademikerin

die Lösungstexte des Gastprofessors

der Antworttext der Englischlehrer

die Bemerkungstexte der Erzieher

der Beschreibungstext der Englischlehrerinnen

die Bemerkungstexte der Akademikerinnen

der Ankündigungstext des Gastprofessors

die Erklärungstexte des Erziehers

der Lösungstext des Dozenten

**Abb. 8:**     Automatisch erhobene Daten aus *Combinatoria*

## 4.    Zum methodologischen Verfahren angesichts eines mehrstufigen Prototypbegriffs

### 4.1    Einführung

Bei der Erstellung der Generatoren (s. 1.) ist einer zentralen methodologischen
Frage nachzugehen, und zwar, wie man semantische Daten automatisch erhebt
und verlinkt, damit sich der Lexemparadigma einer Valenzstelle automatisch
generieren lässt. Erforderlich ist daher eine gründliche Auseinandersetzung
damit, welche Methode bei der Auswahl der vorkommenden lexikalischen Kan-
didaten oder Exemplare zwecks der Belegung der nominalen Stellen (s. Abb. 2)
angewandt wird, ferner wie die semantischen Klassen (s. Abb. 5) festgelegt
werden. Im Konkreten:

— Worauf beruht die Auswahl der lexikalischen Kandidaten der jeweiligen
   semantischen Klassen und wie werden sie den semantischen Rollen und
   Argumentstellen zugeschrieben? Das heißt: Wie wird festgelegt, dass
   'Dozent' eine prototypische Erscheinung bei einer ARG1$_1$ der semantischen
   Klasse »Beruf« beim Substantiv TEXT ist? (s. Abb. 6)

— Worauf begründet sich die Bestimmung einer semantischen Klasse wie
   {ANIMADO HUMANO PROFESIÓN EDUCACIÓN} beim Substantiv TEXT?

Als Ausgangspunkt für weitere Beobachtungen gilt, dass die Festlegung
semantischer Klassen sowie die Auswahl lexikalischer Kandidaten durch mehrere ineinander greifende Beschreibungsprozeduren unterstützt wird. Hiermit stellt sich der den Generatoren zugrunde liegende Prototypbegriff als unentbehrlicher Bestandteil des methodologischen Verfahrens heraus, denn

— erstens erweist er sich bei der Auswahl sowohl der lexikalischen Kandidaten als auch der semantischen Klassen zur Leerstellenbesetzung als eine aussagekräftige Instanz heraus (s. 4.).

— zweitens gilt er als Verlinkungsfaktor mit den WordNet-Ontologien (s. 4.4.).

— drittens findet er auch bei der Bestimmung der biargumentalen Argumentstruktur, der phrasalen Umgebung und des Satzrahmens Anwendung (s. Abb. 2).

## 4.2    Zur Prototypdefinition

Als Prototype fassen wir typische bzw. repräsentative Instanzen für eine konkrete Argumentstruktur oder Slotbesetzung auf. Dabei handelt es sich nicht um semantische abstrakte Konzepte, da der Prototypbegriff mit der syntaktischen Argumentstruktur eines konkreten nominalen Valenzträgers sowie mit der Aktualisierung einer konkreten Bedeutungslesart zusammenhängt (kombinatorische Bedeutung von Engel 2004; vgl. auch die *classes d'objets* bei Gross 2008). Ihre Typikalität sowie Repräsentativität lässt sich auf die Interaktion valenzfundierter und frequenzbezogener Parameter stützen, wie wir noch genauer erläutern werden.

*Per negationem* lassen sich unsere Prototypen weiter näher bestimmen:

— Unsere Prototypen gehen nicht auf kognitionsgebundene Befunde zurück (vgl. Rosch 1978).

— Man ist nicht auf der Suche nach dem besten Exemplar einer semantischen Kategorie, d.h. man versucht nicht festzulegen, ob sich z.B. 'Kopf' als der beste Vertreter einer konkreten Klasse oder Kategorie z. B. {BELEBT MENSCHLICH KÖRPERTEIL} herausstellt. Dementsprechend wird nicht das Ziel angestrebt, die besten oder die schlechteren Vertreter der Kategorie {KÖRPERTEIL} *per se* festzulegen, indem man zentralere und periphere Kategoriebeispiele kontextlos und *per se* unterscheidet.

— Eine angemessene Beschreibung von Wortbedeutungen und darüber hinaus ihre Abgrenzung gegenüber anderen Wörtern bzw. Wortbedeutungen wird von unserem Prototypbegriff nicht geleistet, auch nicht angestrebt.

— Der Prototyp gilt nicht als eine mögliche Alternative zu Definitionen.

— Unsere lexikalischen Prototypen sind keine universellen Kategorien und ihre Festlegung geht nicht auf die Zerlegung ihrer Eigenschaften zurück.

— Die Typikalität bzw. Prototypizität wird nicht gemäß der Ähnlichkeit

eines Kandidaten mit dem "besten" Vertreter einer angegebenen Klasse festgelegt oder beschrieben.

Unterschieden werden in unserem Modell zwei zentrale Prototypenklassen: die lexikalischen Prototypen (4.3.) und die darauf aufbauenden prototypischen semantischen Klassen (4.4.).

### 4.3     Zum Begriff der lexikalischen Prototypen

Um es vorwegzunehmen: Lexikalische Prototypen sind konkrete prototypische Lexeme, die sich bei der Füllung einer bestimmten syntaktisch-semantischen Valenzstelle eines vorgegebenen Valenzträgers als repräsentativ bestimmen lassen. Zu ihrer Festlegung stellt sich die Wechselwirkung von zwei nicht gleichrangigen Kriterien als entscheidend heraus:

— Das ausschlaggebende Kriterium hängt mit der semantischen Belegung einer bestimmten Valenzstelle eines konkreten Valenzträgers zusammen.

— Als untergeordnetes Kriterium gilt die Häufigkeit dieses Kandidats bei Erfüllung einer semantischen Rolle in einem angegebenen phrasalen Rahmen, und dies auch unter Berücksichtigung der Argumentstrukturfrequenz (s. 4.5).

Zur Auswahl der lexikalischen Prototypen muss eingangs die Argumentstruktur des nominalen Valenzträgers beschrieben werden. Dazu lehnen wir uns an dem valenzbasierten Ansatz an, der dem multilingualen Wörterbuch *Portlex* zugrunde liegt (Domínguez Vázquez und Valcárcel Riveiro 2020). Sucht man im Falle vom Substantiv TEXT nach Kandidaten für die Slotbesetzung des Arguments `AGENS´ (`derjenige, der die Handlung durchführt´) realisiert im Muster [TEXT + DETERMINANT IM GENITIV + NOMEN], wird zunächst mittels einer CQL-Abfrage im *Korpus German Web 2013 (de TenTen13)* aus *Sketch Engine* eine Häufigkeitsliste dieser Argumentstruktur erhoben:

|     |                      | Häufigkeit | Häufigkeit pro Million |
|-----|----------------------|------------|------------------------|
| 1.  | Text die Lied        | 1913       | 0.09658                |
| 2.  | Text die Bibel       | 1820       | 0.09188                |
| 3.  | Text die Buch        | 866        | 0.04372                |
| 4.  | Text die Autor       | 844        | 0.04261                |
| 5.  | Text die Band        | 837        | 0.04226                |
| 6.  | Text die Song        | 755        | 0.03812                |
| 7.  | Text die neu Testament | 649      | 0.03276                |
| 8.  | Text die alt Testament | 509      | 0.02570                |
| 9. | Text die Petition | 507 | 0.02560 |
| 10. | Text die Seite | 487 | 0.02459 |
| 11. | Text die Artikel | 467 | 0.02358 |
| 12. | Text die Mail | 365 | 0.01843 |
| 13. | Text die Anzeige | 345 | 0.01742 |
| 14. | Text die Rede | 332 | 0.01676 |
| 15. | Text die heilig Schrift | 331 | 0.01671 |
| 16. | Text die Autorin | 331 | 0.01671 |
| 17. | Text die Urkunde | 326 | 0.01646 |
| 18. | Text die E-Mail | 324 | 0.01636 |
| 19. | Text die Webseite | 323 | 0.01631 |
| 20. | Text die Evangelium | 318 | 0.01605 |

**Tab. 2:**      20 erste Lemmas von TEXT + Genitiv aus dem Korpus *de TenTen13*

Diese Ergebnisse werden dann gemäß semantisch-relationaler Prinzipien herausgefiltert, denn eine formal basierte Häufigkeitsliste reicht für unsere Zwecke nicht aus:

— Bei der Auswahl der lexikalischen Kandidaten besteht Bedarf an einer semantischen valenzbezogenen Analyse im Hinblick auf die relationale Bedeutung[16]. Das geht darauf zurück, dass eine gleiche Form je nach dem Valenzträger und je nach den vom ihm eröffneten Stellen, ebenfalls im Zusammenspiel mit anderen Ergänzungen, unterschiedliche semantische Rollen zum Ausdruck bringen kann. Es liegt nahe, dass die drei ersten häufigsten Beispiele in der Tab. 2 — 'Text des Liedes', 'Text der 'Bibel' und 'Text des Buches' — keine Agensrolle darstellen[17].

— Die bereits angeführten Beispiele bringen ebenfalls den Nachweis dafür, dass die Häufigkeit in unserem Beschreibungsmodell an sich kein allgemeingültiges Kriterium ist. Deutlich wird, dass „frequent" nicht immer mit „valenzbedingt" einhergeht und daher werden 'Text die Monat' (Position: 124; 88 Treffer) und 'Text die Jahr' (Position 103: 102 Treffer) aus der Kandidatenliste ausgeschlossen.

Aus dem Angeführten geht deutlich hervor, dass das Häufigkeitskriterium ausschließlich in enger Verbindung mit der Argumentstruktur, mit der semantischen Rolle und mit der aktualisierten Realisierung und Lesart eines Valenzträgers anzuwenden ist. Aus diesem Grund erstellen wir für jede monolinguale Argumentstruktur jedes Substantivs eine Liste von lexikalischen Kandidaten (grob gesagt, eine Wortschatzliste), die in paradigmatischer Beziehung zueinander stehen. Werden die lexikalischen Kandidaten herausgefiltert, erfolgt ihre kategorielle Annotation in Anlehnung an eine *ad hoc* erstellte Ontologie[18].

Somit stellen wir fest, dass die Stelle `AGENS´ bei einem Substantiv wie TEXT in der Regel von folgenden Substantiven besetzt wird:

| Lexikalischer Prototyp | 1. Stufe | 2.Stufe | 3. Stufe | 4. Stufe |
|---|---|---|---|---|
| Architekt | belebt | menschlich | Beruf | |
| Autor | belebt | menschlich | Beruf | |
| Dichter | belebt | menschlich | Beruf | |
| Hersteller | belebt | menschlich | Beruf | |
| Journalistin | belebt | menschlich | Beruf | |
| Künstler | belebt | menschlich | Beruf | |
| Philosoph | belebt | menschlich | Beruf | |
| Sänger | belebt | menschlich | Beruf | |
| Schriftsteller | belebt | menschlich | Beruf | |
| Schüler | belebt | menschlich | Eigenschaft | |
| Teilnehmer | belebt | menschlich | Eigenschaft | |
| Verlag | belebt | menschlich | Organisation | Unternehmen |
| Uni | belebt | menschlich | Organisation | Bildung |
| EU | belebt | menschlich | Organisation | Politik |
| Paulus | belebt | menschlich | Eigenname | |
| Jury | belebt | menschlich | Kollektiv | |
| Papst | belebt | menschlich | Amt | |

**Abb. 9:**    Kategorielle Beschreibung der lexikalischen Prototypen

Diese Vorgehensweise lässt somit die semantische paradigmatische Belegung einer konkreten Valenzstelle festmachen, die zwar aus sprachwissenschaftlicher Sicht zentral ist, jedoch angesichts der automatischen Sprachgenerierung unerlässlich wird, denn auf die Analyse der prototypischen Kandidaten stützt sich die Festlegung der prototypischen semantischen Klassen. Da der Programmierung der Generatoren nicht einzelne Wörter, sondern semantische Klassen unterliegen, werden sie zum Schlüsselkonzept.

### 4.4    Zum Begriff der prototypischen semantischen Klassen

Bei der Analyse hat sich herausgestellt, dass man sich für die automatische Datenerhebung und Sprachgenerierung (und nicht nur dafür) mit generellen Etiketten wie [Menschen], [Tiere] oder [Gegenstand] nicht zufriedengeben kann, da sie zu weit gefasst sind. Daher erstellen wir auf der Grundlage einer summativen Beschreibung der lexikalischen Kandidaten (Abb. 9) prototypische semantische Klassen. Ein Beispiel dafür bildet die Beschreibung der semantischen Kategorien zur Wiedergabe der Agensrolle bei TEXT (Abb. 10).

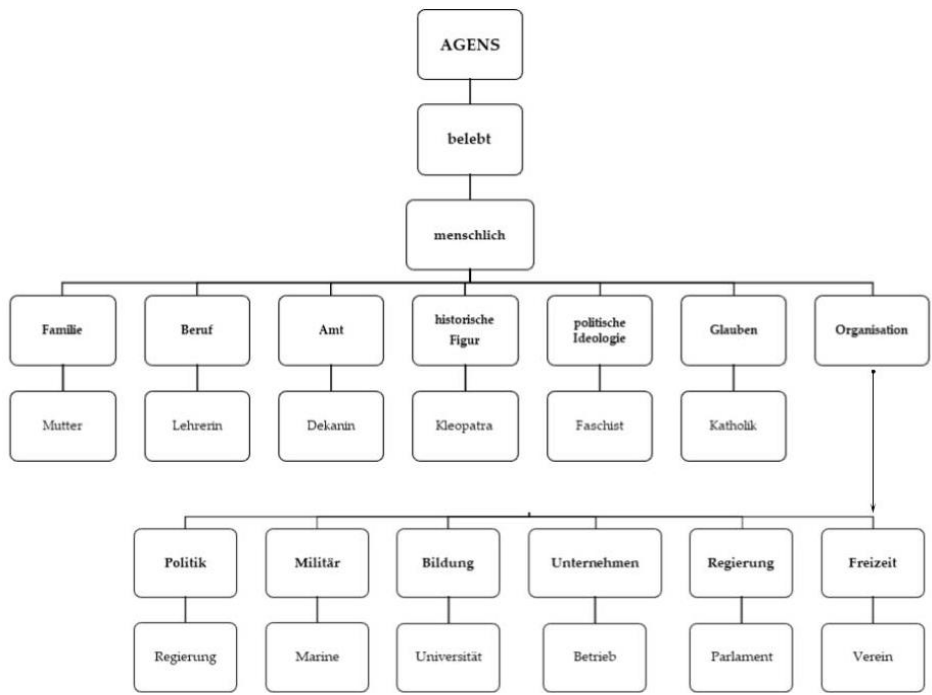

**Abb. 10:**    Semantische Klassen der Agensrolle bei TEXT

Aber damit ist die semantische Analyse nicht ausgeschöpft:

(a) **Zur Beziehung zwischen lexikalischen Kandidaten und semantischen Klassen:** Eine semantische Klasse mit ihren Vertretern lässt sich nicht als allgemeingültige, fixierte Kategorie für jegliche Substantive bestimmen und dementsprechend automatisch verwenden. Zur Verdeutlichung wird die semantische Klasse {BELEBT MENSCHLICH KÖRPERTEIL} herangezogen: Bei der Analyse hat sich erwiesen, dass diese semantische Klasse bei verschiedenen Substantiven, wie z.B. FARBE und SCHMERZ, als prototypisch gilt. Man kann aber dieses lexikalische *Package* — die dieser Klasse zugeordnete Information — nicht automatisch wieder verwenden, da die lexikalischen Kandidaten der semantischen Klasse {BELEBT MENSCHLICH KÖRPERTEIL} bei FARBE und SCHMERZ nicht übereinstimmen[19]:

| Valenzträger | Muster | Lexikalische Prototypen | Frequenz[20] |
|---|---|---|---|
| | | 'Auge' | P.: 31. 460-mal |
| **FARBE** | +[DETERMINANT IM GENITIV+ADJEKTIV[FAKULTATIV]+NOMEN] | 'Haar' | P.: 38. 402-mal |
| | | 'Haut' | P.: 54. 285-mal |
| | | 'Auge' | P.: 30.1949-mal |
| **SCHMERZ** | +[DETERMINANT + {KÖRPERTEIL}- +SCHMERZ] | 'Haar' | P.: 263. 39-mal |
| | | 'Haut' | P.: 234. 48-mal |

**Tab. 3:**     Gegenüberstellung der Kandidaten einer semantischen Klasse bei verschiedenen Nomina. Daten aus *Sketch Engine*

Bei SCHMERZ lassen sich 'Haar' und 'Haut' im Gegenteil zu FARBE nicht als prototypische Vertreter dieser semantischen Klasse auffassen, vor allem wenn man ihre Ko-Okurrenzen den prototypischen und häufig vorkommenden Komposita gegenüberstellt (Tab. 4).

| | | 'Kopfschmerz' | P.:1. 188948-mal |
|---|---|---|---|
| **SCHMERZ** | +[DETERMINANT + {KÖRPERTEIL}- +SCHMERZ] | 'Rückenschmerz' | P.:2. 92363-mal |
| | | 'Bauchschmerz' | P.: 3. 45907-mal |

**Tab. 4:**     Lexikalische Prototypen bei SCHMERZ. Daten aus *Sketch Engine*

(b) **Zur Wortschatzerweiterung:** Bezüglich der Erstellung der Generatoren ist einer weiteren Frage nachzugehen, die auch an die Wichtigkeit der Festlegung semantischer prototypischer Klassen anknüpft: die Rolle der semantischen Klassen oder Kategorien, und ebenfalls der konkreten kategoriellen Merkmale ([Menschen], [Organisation]), als Verknüpfungsinstanz mit Kategorien und Sub-kategorien der WordNet-Ontologien (Gómez Guinovart und Solla Portela 2018). Dabei sind die semantischen Klassen von herausragender Bedeutung, und dies aus den folgenden Gründen:

— Die Verlinkung auf die WordNet-Ontologien mithilfe der semantischen Klassen ermöglicht eine verfeinerte Granularität unserer Ontologie sowie eine detailliertere Beschreibung des sprachlichen Materials.

— Dadurch gelangt man zu einer vollständigeren Wortschatzliste für jede konkrete Leerstellenbesetzung.

Dieses Verfahren der Wortschatzerweiterung trägt zur Bildung eines selegierten Lexemparadigmas mit den gleichen semantischen Eigenschaften wie denen des ursprünglichen lexikalisch-semantischen Prototyps bei. So wird die paradigmatische Achse festgelegt, die die lexikalische Auswahl bei der automatisierten Generierung von Nominalphrasen unterstützt.

Die Expandierung erfolgt dank der Anwendung der Tools *Lematiza*[21] und *Combina*[22] (Domínguez Vázquez, Solla Portela und Valcárcel Riveiro 2019). Durch ihren Einsatz können wir Wortschatzlisten aus den WordNet-Ontologien — z.B. {KÖRPERTEIL} in der Tab. 5 — gezielt zusammenstellen, aber auch semantische Klassen genauer einteilen (Tab. 6). Diese Einteilung in Klassen und Unterklassen ist enorm relevant: denn — nochmals am Beispiel vom {KÖRPERTEIL} — wir alle wissen, dass nicht alle Körperteile riechen oder wehtun, d.h. nicht alle Körperteile kommen im gleichen Prädikatausdruck vor.

| WordNet: Ontologien[23] | Klassen |
|---|---|
| *SUMO Ontology* | BodyPart+ BodyJunction+ Organ + |
| *Epinonyms* | [1] external_body_part |
| *Basic Level Concept* | 05220461-n body_part |
| *WordNet Domains* | anatomy |
| *Top Concept Ontology* | 1stOrderEntity+ Living+ Part+ |

**Tab. 5:**    {KÖRPERTEIL} bei WordNet-Ontologien

| Klassen | Vertreterbeispiel |
|---|---|
| [belebt] [menschlich] [Körperteil] [extern] | 'Kopf' |
| [belebt] [menschlich] [Körperteil] [intern] [Muskel/Knochen] | 'Muskel' |
| [belebt] [menschlich] [Körperteil] [Beschichtung] | 'Haut' |
| [belebt] [menschlich] [Körperteil] [Organ] | 'Eierstock' |

**Tab. 6:**    Semantische Klassen mit {KÖRPERTEIL}

Bei diesem Verfahren steht der Gedanke einer gesteuerten Erhebung von einer signifikativen Anzahl an lexikalischen Kandidaten im Vordergrund. Dieser zuletzt genannte Aspekt ist nicht zu unterschätzen, denn bei der qualitativen Evaluation von automatisch generierten Daten (Hashimoto et al. 2019, Vicente et al. 2015) sind nicht nur ihre Qualität, sondern auch ihre Vielfaltigkeit zentral.

## 4.5    Prototypen als extern und intern vergleichende Instanzen

### 4.5.1    Einführung

Bei der praktischen Anwendung unseres Prototypizitätbegriffs bzw. Typikalitätsbegriffs sind wir auf unterschiedliche Fälle gestoßen, die einer näheren Beobachung bedürfen und die Prototype näher bestimmen lassen:

— Prototyp als **extern vergleichende Instanz:** Diese Auffassung ermöglicht die Gegenüberstellung eines konkreten lexikalischen Kandidats bei unterschiedlichen nominalen Valenzträgern. Ein Beispiel dafür bildet der schon angeführte Vergleich von {KÖRPERTEIL} bei FARBE und SCHMERZ (vgl. 4.4.). Eine weitere Anwendung des Prototyps als extern vergleichende Instanz kann auch bei der Gegenüberstellung verschiedener Sprachen erfolgen. Dies lässt sich beim Vorkommen der Eigennamen in den 3 beschriebenen Sprachen deutlich beobachten. Somit sind prototypisch 'Mia' oder 'Lena' im Deutschen, im Spanischen 'Susana' oder 'Patricia' und im Französischen 'Paulette' oder 'Annick'.

— Prototyp als **intern vergleichende Instanz:** Zu einem internen Vergleich erweist sich ebenfalls der Prototyp als ausschlaggebend, ferner bestätigt dieser ihn als solchen. In diesem Zusammenhang ist dem Verhältnis zwischen der Repräsentativität der Argumentstruktur, der Anzahl an lexikalischen Kandidaten sowie der Häufigkeit der lexikalischen Kookurrenzen ein zentraler Stellenwert zuzuweisen. Darauf wird in den nachstehenden Abschnitten eingegangen.

### 4.5.2   Prototypen: Argumentstruktur und Kandidantenliste

Als intern vergleichende Instanzen lassen sich die Prototypen nach verschiedenartigen Prototypizitätstypen oder -grade auf einer fiktiven Prototypizitätskala unterscheiden:

(i)   Als prototypischste Argumentstruktur stellt sich ein repräsentatives und häufig vorkommendes Muster mit einer breiten Palette an verschiedenen lexikalischen Kandidaten heraus, die außerdem häufig vorkommen. Ein Beispiel dafür bildet die Argumentstruktur [DISKUSSION + Präposition^(über/um) + Nomen] bei Wiedergabe der semantischen Rolle `THEMA´:

|  | Häufigkeit | *Scores* |
|---|---|---|
| Diskussion über | **114,929** | **6.57%** |
| Diskussion um | 110,558 | 6.32% |
| Diskussion zu | 47,816 | 2.73% |
| "Diskussion" außerhalb + noun | 21 | 0,0% |
| "Diskussion" anlässlich + noun | 15 | 0,0% |
| Diskussion bzgl. | 13 | 0,0% |

**Tab. 7:**   Gegenüberstellung von Frequenzangaben bei DISKUSSION + Präposition

Die Tabelle gibt zu erkennen, dass das Muster [DISKUSSION + Präposition^(über) + Nomen] häufiger als andere Kombinationen vorkommt und dass es im Vergleich zu anderen möglichen Realisierungsmustern, z.B. [Komposita + -DISKUSSION], bei Wiedergabe derselben semantischen Rolle repräsentativ ist. Hinzu kommt, dass sie über unterschiedliche und häufig vorkommende lexikalische Kandidaten verfügt:

| *Diskussion über* | Frequenz |
|---|---|
| Thema | 5,189 |
| Zukunft | 2,886 |
| Sinn | 1,893 |
| Frage | 2,383 |
| Rolle | 874 |
| Problem | 874 |
| Einführung | 714 |
| Umgang | 743 |
| Inhalt | 765 |
| Möglichkeit | 838 |

**Tab. 8:**     Lexikalische Kandidaten nach Häufigkeitsangaben

(ii)   Im Gegensatz zu (i) sind zwar nicht häufige Argumentmuster vorhanden, die sich aber als repräsentativ herausstellen: Aus einer valenzbasierten Analyse der aus *Sketch Engine* erhobenen Daten über das Muster [Adjektiv + SCHMERZ] lässt sich ableiten[24], dass nicht valenzbezogene Realisierungen bezüglich der Gesamtanzahl der Adjektive als auch der Ko-Okurrenzen in der Mehrheit stehen. Somit zeigt eine semantisch basierte Analyse der ersten 100 Substantive, dass 92% der vorkommenden Adjektive nicht valenzgefordert sind, bei den restlichen 8% handelt es sich um Ergänzungen.

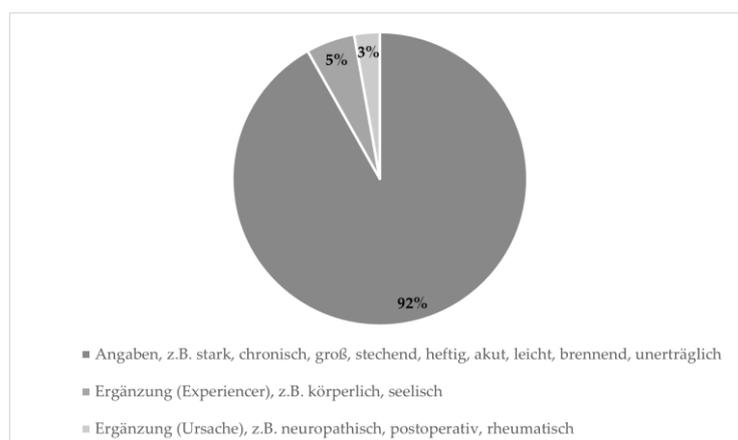

**Graphik 1:**     Semantisch valenzbasierte Analyse von Adjektiven bei SCHMERZ

In Bezug auf die Häufigkeit der valenzbezogenen Realisierungen lässt sich das Muster zweifelsohne nicht als häufig bezeichnen. Indem es sich aber als repräsentativ für eine mögliche Realisierung der semantischen Rolle `EXPERIENCER´ erweist, wird es in den Generatoren berücksichtigt.

Hiermit lässt sich ebenfalls beobachten, dass dieses Muster eine sehr eingeschränkte Kandidatenliste zeigt, deren Vertreter — 'körperlich' (5,899 Treffer; 5. Häufigkeitsposition) und 'seelisch' (10. Häufigkeitsposition mit 3,621 Treffer) — aber ganz oben in der Frequenzliste stehen.

|   |                  | Häufigkeit | Häufigkeit pro Million |
|---|------------------|------------|------------------------|
| 1 | stark Schmerz    | 30,125     | 1.52                   |
| 2 | chronisch Schmerz| 21,934     | 1.11                   |
| 3 | groß Schmerz     | 12,875     | 0.65                   |
| 4 | stechend Schmerz | 8,063      | 0.41                   |
| 5 | körperlich Schmerz| 5,899     | 0.30                   |
| 6 | heftig Schmerz   | 5,165      | 0.26                   |
| 7 | akut Schmerz     | 5,045      | 0.25                   |
| 8 | leicht Schmerz   | 4,624      | 0.23                   |
| 9 | brennend Schmerz | 3,673      | 0.19                   |
| 10| seelisch Schmerz | 3,612      | 0.18                   |

**Tab. 9:** Häufigkeitsangaben von Adjektiven + SCHMERZ

(iii) Nicht repräsentative wie auch nicht häufig vorkommende Argumentstrukturen mit wenigen und nicht frequenten Kandidaten werden nicht in die Generatoren integriert. Zur Veranschaulichung derartiger Fälle dient das Muster [SCHMERZ + Genitiv: {BELEBT MENSCHLICH KÖRPERTEIL}]. Hingegen deuten die Korporadaten zur Prototypizität der Kompositarealisierung hin:

|                      | Häufigkeit | Häufigkeit pro Million |
|----------------------|------------|------------------------|
| 107 Schmerz die Kopf | 21         | < 0.01                 |
| 1 Kopfschmerz        | 188950     | 954                    |

**Tab. 10:** Gegenüberstellung von 'Kopf' in Verbindung mit SCHMERZ

### 4.5.3 Prototypen: semantische Klassen und Argumentstruktur

Hinsichtlich des Verhältnisses zwischen semantischen Klassen und Argumentmustern sind zwei Fälle zu beachten:

(i)     Integriert werden in die Generatoren semantische Klassen mit einer signi-
        fikativen Anzahl an Vertretern, die sich einer häufigen und repräsentati-
        ven Argumentstrukturmuster zuschreiben lassen. Als Beispiel stellt sich
        die semantische Klasse {BELEBT MENSCHLICH KÖRPERTEIL} in Form eines
        Kompositumsbestandteils bei SCHMERZ heraus:

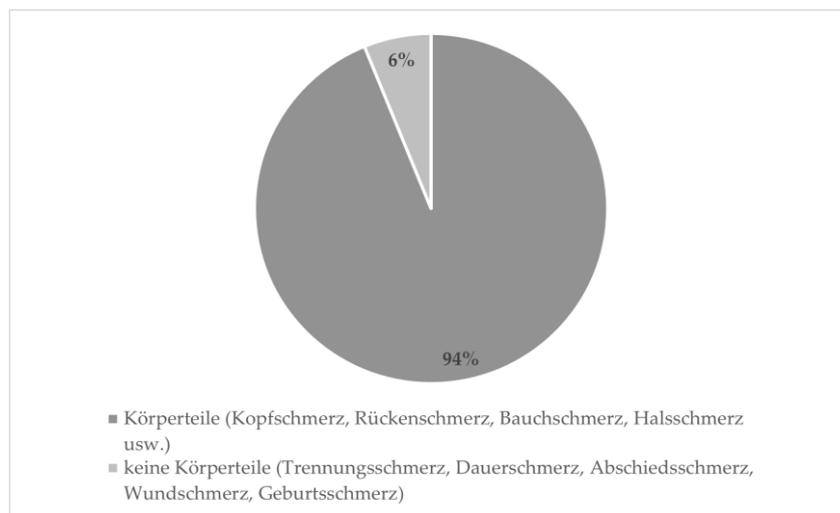

**Graphik 2:**     Prozentansätze von Komposita mit Körperteil bei SCHMERZ

(ii)    Im Gegensatz zu (i) stößt man auch auf Fälle von semantischen Klassen
        mit wenigen aber häufig vorkommenden Vertretern. Ein Beispiel dafür
        findet das syntaktische Muster [FARBE + Genitiv], dessen Genitivrealisie-
        rung mit vielen semantischen Klassen ausgestattet ist. Darunter lässt sich
        als Beispiel dafür die semantische Klasse {MATERIELL GEGENSTAND
        SCHÖNHEITSPLEGE KOSMETIK} nennen, die eine eingeschränkte Anzahl
        an Vertretern hat — z.B. 'Lippenstift' — , die aber schon sehr häufig auf-
        tauchen.

### 4.5.4   Prototypen in den biargumentalen Strukturen

Bisher war von monoargumentalen Strukturen die Rede. Deutlich werden
sollte allerdings unmittelbar, dass sich bei der Entfaltung des im Nomen enhal-
tenen Potentials nicht alle Kombinationen und Selektionsbeschränkungen vor-
aussehen bzw. voraussetzen lassen. Im Hinblick darauf liegt es nahe, dass nicht
alle Vertreter jeder semantischen Klasse eines Arguments[X] mit allen Vetretern
einer anderen Klasse eines Arguments[Y] in einer biargumentalen Realisierung
kombinierbar sind.

Bei der Festlegung biargumentaler Muster greift man wieder auf den Prototypbegriff zurück. Somit ist es erforderlich, zuerst das Vorhandensein des biargumentalen Musters im Korpus zu prüfen. Liegt der Analyse das Muster [Determinant + {Adjektiv} + ANTWORT + Determinant im Genitiv + Argument N1 + auf + Determinant im Akkusativ + Argument N2] zugrunde, lässt sich festhalten, dass die semantische Klasse {ORGANISATION REGIERUNGSGEBUNDEN} für N1 und die semantische Klasse {INTELLEKTUELLES KOMMUNIKATION} für N2 prototypisch sind. Beachtet man konkret die vorkommenden lexikalischen Prototypen, ist zu beobachten, dass die lexikalischen Kandidaten 'Anfrage' und 'Frage' in der Mehrheit stehen:

| | | |
|---|---|---:|
| | die Bundesregierung auf eine Anfrage | 465 |
| | die Landesregierung auf eine Anfrage | 92 |
| | die Senat auf eine Anfrage | 81 |
| | die Bundesregierung auf die Anfrage | 66 |
| ANTWORT | die Landesregierung auf die Anfrage | 38 |
| | die Regierung auf eine Anfrage | 37 |
| | die Verwaltung auf eine Anfrage | 35 |
| | die Bundesregierung auf die Frage | 35 |
| | die Stadtverwaltung auf eine Anfrage | 31 |
| | die Verwaltung auf die Anfrage | 29 |

**Tab. 11:**    Biargumentales Muster von ANTWORT

Zum Erlangen akzeptabler Ergebnisse bei der sprachlichen Generierung ist zwar eine vertiefte Analyse der biargumentalen Muster unentbehrlich, jedoch kann jede mögliche semantisch-syntaktische Kombination nicht manuell analysiert werden. Aus diesem Grund lassen sich die Resultate unseres kombinatorischen Tools — *Combinatoria* — mithilfe der prädiktiven Methode *Word2vec* (Mikolov et al. 2013) auf die Kompatibilität der lexikalischen Kandidaten jedes Arguments überprüfen (Domínguez Vázquez 2020). Angestrebt wird dabei nicht, die Bedeutung eines Wortes einem anderen gegenüberzustellen — wie beim *Embedding Viewer* von *Sketch Engine* oder die möglich abrufbar paradigmatische Datenerhebung aus *Derekovecs*[25], sondern die Beseitigung von Inkompatibilitäten auf der Ebene der kombinatorischen Bedeutung. Dieses Vorgehen trägt zur semantischen Kohärenz bei, einem Qualitätsfaktor bei jedem Generator natürlicher Sprachen. Der Zugriff auf *word embeddings* wird auch die Integration von nicht valenzbezogenen Adjektiven in die nominalen Argumentmuster ermöglichen, damit der phrasale Kontext vermittelt wird: *Der {unangenehme} | {angenehme} | {üble} | {intensive} Geruch.*

## 5.     Auf dem Weg zum automatischen Wörterbuch

Bei den vorangehenden Abschnitten ist auf die Integration und Rückkoppe-
lung verschiedener Arbeitsschritte und Beschreibungsebenen zwecks der auto-
matisierten Datenerhebung und -generierung von syntaktisch-semantischen
Valenzmustern eingegangen. Der Fokus hat insbesondere auf der Entwicklung
eines mehrsichtigen Prototypbegriffs gelegen, denn dieser stellt sich für die
Verlinkung zwischen zwei zentralen Beschreibungsebenen als ausschlaggebend
heraus: zum einem die automatische Erhebung semantischer Daten, zum anderen
die Generierung des syntaktisch-semantischen Entfaltungspotentials der Nomi-
nalphrase.

Die Sprachgeneratoren, *Xera* und *Combinatoria*, können als selbständige
Tools, als lexikalische bzw. sprachliche Informationssysteme (s. Villa Vigoni-
Thesen 2018) gebraucht werden. Darüber hinaus fassen wir sie als innovative
Modelle für automatische und interaktivere Valenzwörterbücher bzw. syntag-
matische Kombinationswörterbücher auf. Sie bringen gegenüber von aktuellen
online syntagmatischen Wörterbüchern gewisse Neuerungen mit sich:

Die semantischen Rollen, die prototypischen semantischen Klassen und
die lexikalischen Prototypen gelten bei der Erstellung und bei der Hanhabung
der Ressource als entscheidende Bausteine. Insofern bieten die Generatoren
nicht nur eine ontologisch fundierte Beschreibung, sondern auch ein Inventar
an semantischen Kategorien und an lexikalischen Kandidaten zur Belegung
einer konkreten Valenzstelle an. Somit leisten sie, wie klassische Valenzwörter-
bücher, Hilfe in einer typischen Produktionssituation.

Aus typologischer Sicht bestimmen zwei Schlüsselkonzepte, automatisch
und interaktiv, den aktuellen Stand der Prototypen:

—     Es handelt sich um automatische Ressourcen in dem Sinne, dass sowohl
        für die Datenerhebung als auch für die Datengenerierung Techniken und
        Verfahren aus der NLP und NLG eingesetzt bzw. neu konzipiert wurden.
        Die Generatoren können eine Grundlage für die Entwicklung künftiger
        automatischer bzw. automatisch computergestützer Valenzwörterbücher
        bilden.

—     Beide Ressourcen sind ebenfalls als interaktive Tools konzipiert, indem im
        Gegenteil zu früheren syntagmatischen Werken eine Interaktionshand-
        lung zwischen Werk und Nutzer sttatfindet. Bei ihrer Handhabung kann
        in ihrem aktuellen Stand eine informativ-beschreibende Interaktion erfolgen.
        Folglich kann man zur Vergewisserung in einer Produktions- oder Korrek-
        tursituation gezielt bestimmte Muster bzw. Kombinationen abfragen.
        Zwei interaktive Herangehensweisen befinden sich in einer Entwicklungs-
        phase: (i) eine experimentativ-nachschlagende Interaktion (Selbstgenerie-
        rung und -überprüfung eines Ausdrucks) und (ii) eine experimentativ–
        betreute Interaktion (Durchführung der von den Tools angebotenen online
        Übungen). Insgesamt sind die didaktischen Anwendungen der Generato-

ren vielfältig (Domínguez Vázquez, Sanmarco Bande, Solla Portela und Valcárcel Riveiro 2019: 133-135).

Neben der Anwendung der Generatoren als selbständige Ressourcen und als Modelle für neuartige Valenzwörterbücher ist ein weiterer Anwendungsbereich zu ergänzen: die Integration ihrer Daten in andere E-Tools sowie die Anwendung der der Entwicklung der Generatoren zugrundeliegenden Tools für andere Informationssysteme. Beide sind wichtige Aufgaben der elektronischen Lexikographie:

> Another aspect is smart use and reuse of dictionary content. Namely, dictionaries often remain isolated entities, whereas the user needs and habits indicate that it would be much more useful to have them linked to other dictionaries and language resources, or even integrated in various tools. (siehe Elex 2019[26])

## 6.      Ausblick

Eine ausführliche Analyse der syntaktisch-semantischen Schnittstelle ist aus linguistischer Perspektive zentral, da die konkrete Besetzung der vom Valenzträger eröffneten Stellen zur Prädikatsinterpretation und darüber hinaus zur Abgrenzung gegenüber anderen Prädikaten führt. Aus Programmierungssicht ist ebenfalls die semantische Information unentbehrlich, denn Maschinen verfügen nicht über ein semantisches Wissen, daher findet in unserem Modell eine syntaktisch-semantisch valenzfundierte Beschreibung des vom Valenzträger festgelegten syntaktisch-semantischen Rahmens statt. Im Hinblick auf die notwendige syntagmatische und paradigmatische Selektion der von Generatoren gelieferten Daten, ist ein mehrstufiger Prototypbegriff entwickelt worden, der sich als wesentlicher Bestandteil des methologischen Verfahrens erweist. Sowohl aus der Entwicklungsperspektive als auch aus der Nutzersicht rückt in unserem Ansatz die Darlegung der kombinatorischen Bedeutung auf phrasaler Ebene in den Vordergrund. Die automatische Generierung von satzwertigen Realisierungen stellt sich als nächstes Ziel heraus.

Beide schon einsatzbereite Prototype verkörpern einen typologischen Vorschlag zu neuen automatischen computergestützten syntagmatischen pluringualen Wörterbüchern, die ebenfalls unterschiedliche Interaktionen beachtet.

Insgesamt zeichnet der Beitrag nach, wie neue lexikographische Tools auf der Grundlage des schon vorhandenen lexikograpischen Wissens, der Interaktion zwischen der Lexikographie und der natürlichen Sprachverarbeitung (NLP) und -generierung (NLG) sowie der Datenintegration und Ressourceninteroperabilität zustande kommen können.

## Danksagung

Die Ergebnisse dieser Forschung stehen im Zusammenhang mit dem Untersuchungsvorhaben „Mehrsprachige Generierung von nominalen Argumentstrukturen und automatische syntaktisch-semantische Datenerhebung", gefördert vom Programm „Ayudas Fundación BBVA a Equipos de Investigación Científica 2017", sowie mit dem Forschungsprojekt „Mehrsprachige Generierung von nominalen Argumentstrukturen mit Anwendung bei der Produktion in Fremdsprachen", gefördert von FEDER/spanischem Ministerium für Wirtschaft, Industrie und Wettbewerb — Staatliche Forschungsagentur (FFI2017–82454-P; Programm Exzellenz).

## Endnoten

1.   Die verwendeten Korpora bei der Erstellung dieses Wörterbuchs sind: für Deutsch *COSMAS II* (https://www.ids-mannheim.de/cosmas2/web-app/), für Französisch *FRANTEXT* (https://www.frantext.fr/), für Galicisch *CORGA* (http://corpus.cirp.gal/corga), für Italienisch *PAISÀ* (https://www.corpusitaliano.it/) und für Spanisch *CREA* (http://corpus.rae.es/creanet.html). In einigen Fällen haben wir auf *Sketch Engine* oder das Web als Korpus zugegriffen.

2.   Ressourcen, die syntaktisch-semantische Informationen anbieten, stehen zur Verfügung. Ein Beispiel dafür bilden *Propbank*, *CPA* (Hanks und Pustejovsky 2005) oder *AnCora*.

3.   Sie sind unter dem Link http://portlex.usc.gal/combinatoria abrufbar.

4.   Siehe http://adimen.si.ehu.es/web/MCR.

5.   Andere Autoren wie Kilgarriff et al. (2008) haben sich mit der Findung guter Wörterbuchbeispiele im Korpus auseinandergesetzt. Hervorzuheben sei hier, dass unsere Belege das valenzfundierte Entfaltungspotential und seine Varietät aufzeigen müssen, nur angesichts der Ressourcetypologie lassen sich dann als gute Beispiele bezeichnen.

6.   Alle Frequenz- und Korpusdaten, auf die man sich im Beitrag bezieht, sind den Korpora aus *Sketch Engine* entnommen. Im Konkreten sind die Daten für das Deutsche dem *Korpus German Web 2013 (de TenTen13)* entnommen.

7.   Im Konkreten: *Sumo Ontologie* (Niles und Pease 2001), *Epinomys* (Gómez Guinovart und Solla Portela 2018), *Basic Level Concept* (Izquierdo et al. 2007), *WordNet Domains* (Bentivogli et al. 2004) und *Top Concept Ontology* (Álvez et al. 2008).

8.   Bei Arbeiten wie Renau und Nazar (2016) geht es um die automatische Mustererhebung (*pattern*), nicht um die Generierung.

9.   In der Graphik wird auf unterschiedliche Ressourcen und *Tools* Bezug genommen, die bei den verschiedenen Entwicklungsphasen der Generatoren verwendet werden. Voneinander abzugrenzen sind bereits vorhandene Tools — wie das Korpustool *Sketch Engine*, das *Freeling Tagger* (Padró et al. 2010) oder semantische Netze, wie *WordNet* (Gómez Guinovart und Solla Portela 2018) — gegenüber denen, die für das Forschungsvorhaben *ad hoc* entwickelt wurden, — wie *APIs* und die Tools *Lematiza*, *Combina*, *Flexiona* (siehe dazu Domínguez Vázquez, Solla Portela und Valcárcel Riveiro 2019). Frei zugänglich sind alle unsere Ressourcen bis auf eine Ausnahme: der Generator des Satzrahmens, der gerade getestet wird.

10.  Für weitere Beschreibungsphasen siehe 4.

11.  Da die Generatoren ebenfalls Daten fürs Spanische und Französische anbieten, sind ebenfalls die Argumentstellen in diesen Sprachen einzeln beschrieben. Für weitere Details siehe Domínguez Vázquez und Valcárcel Riveiro (2020) und Valcárcel (2017: 193). Auflösung der Abkürzungen: GS steht für Genitivus Subiectivus, PrpE für Präpositivergänzung und NmlE für Nominalergänzung.

12.  Die in der Tabelle angeführten kategoriellen Beschreibungskategorien werden bei der Entwicklung der Prototypen weiter spezifischer differenziert, hier gelten sie als Ausgangspunkt. Vgl. dazu auch Martín Gascueña und Sanmarco Bande (2019).

13.  Es besteht die Möglichkeit, die Daten einer bestimmten semantischen Klasse, unterschiedlicher Klassen oder aller Klassen gleichzeitig abzurufen.

14.  Übersetzung im Deutschen: {BELEBT MENSCHLICH BERUF AUSBILDUNG}.

15.  Da die Selektion der Argumente im Auswahlmenü für Laien Schwierigkeiten bereiten kann, werden zur Zeit andere Zugriffstrukturen getestet. Ein Beispiel dafür bietet den Vorschlag zu einem ontologisch-semantischen Datenzugriff im Tool *XeraWord* (Domínguez Vázquez et al. 2020), das auf der Grundlage von *Xera* (Domínguez Vázquez et al. 2020) für das Galicische und Portugiesische entwickelt worden ist.

16.  Eine semantisch verankerte Datenerhebung ist bei den verwendeten Korpora — wie bei vielen anderen — nicht möglich, da sie nicht semantisch annotiert sind.

17.  Es handelt sich eigentlich um das Argument — `dasjenige, das etwas hat/über etwas verfügt´ —, zwar eine Ergänzung, aber kein `AGENS´ (vgl. Abb. 2).

18.  Die erstellte Ontologie vereinbart vorhandene Inventare der Valenzlexikographie, Daten aus WordNet sowie ein *bottom-up* -Verfahren.

19.  Da die einer semantischen Kategorie zugeschriebene Kandidatenliste bei allen beschriebenen Substantiven nicht gleich sein muss bzw. nicht ist, müssen die erstellten Kandidantenlisten bei jedem Substantiv geprüft werden.

20.  P. steht für Position.

21.  Die Lexeme, die wir aus den Korpora in *Sketch Engine* erhoben haben, bietet *Lematiza* begleitet von ihren jeweiligen *synsets* und verlinkt auf die verschiedenen WordNet-Ontologien automatisch an.

22.  *Combina* ermöglicht kombinierte Abfragen an die WordNet-Ontologien sowie die gezielte Erhebung lexikalischer Kandidaten mit ihren *synsets*.

23.  Zu den WordNet-Ontologien siehe Fußnote 7.

24.  Es herrscht kein Konsens darüber, ob Adjektive sowie Kompositabestandteile als Ergänzungen fungieren können. Hier wird der Ansatz vertreten, dass die syntaktisch-semantische Funktion nicht von der Form bestimmt wird und dementsprechend diese Formen auch die Funktion einer Ergänzung erfüllen können.

25.  Unter: http://corpora.ids-mannheim.de/openlab/derekovecs/.

26.  Unter: https://elex.link/elex2019/.

## Literaturverzeichnis

## Online-Ressourcen